# Generative Animations: A Multi-Model Pipeline for Prompt-Driven Motion Synthesis

Mannat Khurana, Sanyam Jain, Rishav Agarwal

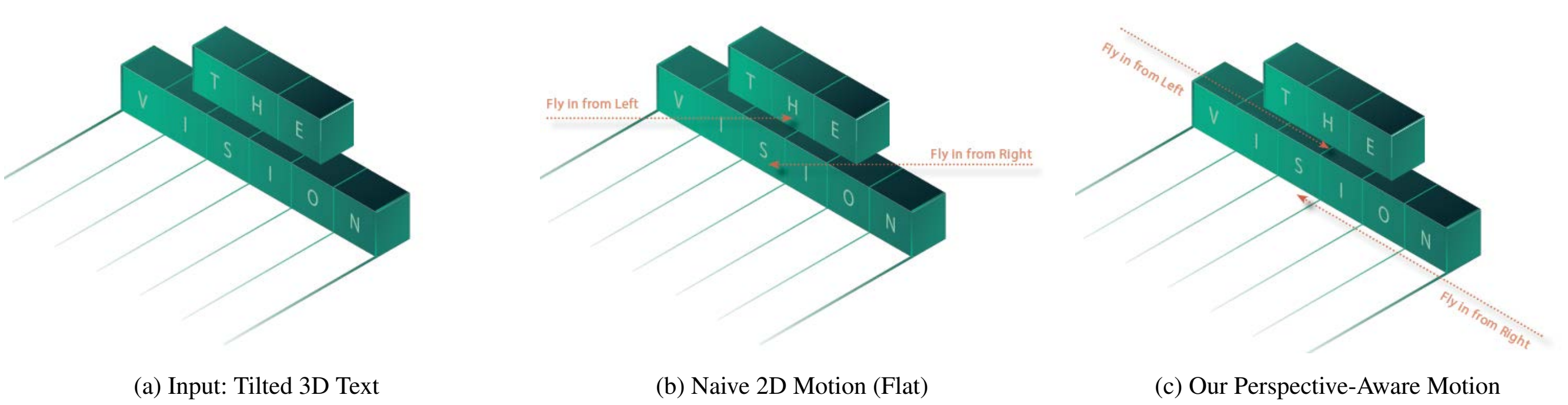

(a) Input: Tilted 3D Text (b) Naive 2D Motion (Flat) (c) Our Perspective-Aware Motion

Figure 1: Generative Animations automatically project motion paths into the 3D coordinate space of the subject, ensuring geometrically correct results that respect the existing document layout.

**Abstract**

*Animation elevates digital documents into immersive experiences, yet creating custom motion paths remains cumbersome, requiring designers to manually select presets, plot Bézier points, and configure timing properties. We introduce* Generative Animations, *a system that transforms natural language prompts into production-ready animations. By chaining Large Language Models (LLMs) for semantic parsing with the Segment Anything Model (SAM) for visual grounding, our pipeline automatically generates motion paths that respect scene geometry, handle depth-based occlusions, and honor 3D perspective transforms. We demonstrate the system through three use cases: contour-following trajectories, orbital animations with z-order awareness, and perspective-aligned motion on transformed objects.*

## 1. Introduction

Animation transforms static digital content into dynamic, engaging experiences and has become essential across industries: marketing teams use animated social media posts to capture attention, educators employ motion graphics to explain complex concepts, and e-commerce platforms leverage product animations to drive conversions. The global motion graphics market continues to expand as businesses recognize animation's power to communicate and persuade.

Despite this demand, creating custom animations remains a significant b ottleneck. P rofessional t ools l ike A dobe A fter Effects and Apple Motion offer powerful capabilities but require extensive training and manual effort. Even in streamlined applications like Canva or Adobe Express, users face a multi-step process: selecting from limited presets, manually plotting Bézier control points with the Pen Tool, and configuring timing curves. This complexity creates a gap between creative vision and execution, particularly for non-specialist users who lack motion design expertise (see Figure 1).

Consider animating a character walking along a curved terrain in a landscape illustration. A designer must: (1) visually identify the path contour, (2) manually trace it point-by-point, (3) convert the trace to a motion path, and (4) configure timing and easing. For a moderately complex path, this process requires 5-10 minutes of focused work per animation. Multiply this across a typical project with dozens of animated elements, and the time investment becomes prohibitive.

**Our Contribution.** We introduce *Generative Animations*, an intelligent system that bridges this gap by translating natural language prompts into production-ready animations. With a command as simple as "move Mario along the hilly path," our system automatically: (1) parses intent using an LLM, (2) grounds the reference to visual elements via SAM [KMR*23], (3) synthesizes smooth motion paths, and (4) assembles the final animation. Our key contributions are:

- A **multi-model pipeline** combining LLM-based semantic parsing with SAM-based visual grounding to translate prompts into structured animation specifications.
- **Environment-aware path synthesis** that generates motion trajectories respecting scene contours, depth ordering, and 3D perspective transforms.
- **Broad applicability** demonstrated through integration with Adobe InDesign and Express, achieving 90% time savings across diverse animation scenarios.

## 2. Related Work

**LLMs for Creative Applications.** Large Language Models [Ope23, BMR*20] have enabled natural language interfaces across creative domains, from code generation to image editing. Recent work demonstrates LLMs orchestrating multi-step creative workflows by decomposing high-level instructions into tool invocations [WYQ*23]. Our work applies LLMs to animation authoring, using them for *semantic parsing* of user intent into structured animation specifications.

**Visual Foundation Models.** SAM [KMR*23] provides zero-shot segmentation, enabling precise object isolation without task-specific training. Grounding DINO [LZR*24] extends this with text-conditioned detection. We leverage SAM for geometric grounding, extracting motion-relevant contours from document artwork rather than editing static images.

**Text-to-Motion Generation.** Recent advances in human motion synthesis, such as MDM [TRG*23] and MotionGPT [JCL*23], generate realistic 3D character animations from text. However, these focus on skeletal motion for 3D characters. Our work addresses a complementary problem: generating 2D motion paths for arbitrary graphic elements within document layouts.

**Animation Authoring Tools.** Professional tools (After Effects, Motion) require extensive manual effort. Lightweight formats like Lottie [Air17] enable cross-platform playback but still require manual authoring. Our approach automates the *path specification* step while leveraging existing interpolation and playback engines.

## 3. System Architecture

Our system follows a unified four-stage pipeline that transforms natural language into environment-aware animations (Figure 2). Each stage addresses a distinct challenge in bridging intent to execution.

**Stage 1: Semantic Parsing.** A GPT-4 class LLM receives the user's prompt along with a system prompt that encodes: (a) available animation presets (Fade, Fly-In, Rotate, Bounce, Gallop, Wave, etc.), (b) expected JSON output schema, and (c) domain context about design applications. We utilize *few-shot prompting* with example pairs of prompts and JSON outputs to steer the model towards syntactically valid and semantically accurate extractions. The LLM extracts structured intent including the animation subject, reference entity, preset selection, and timing parameters.

**Stage 2: Visual Grounding.** The extracted entity name is passed to SAM [KMR*23] to generate a pixel-accurate mask. Our system implements a *multi-candidate resolution strategy*: if the document contains multiple paths or objects matching the LLM's entity description, the system presents the candidate masks to the user. The user provides a single point-click for disambiguation, which is used as a positive prompt for SAM to refine the selection. This maintains a minimal interaction overhead while ensuring zero-error grounding in cluttered scenes.

**Stage 3: Path Synthesis.** The segmentation mask undergoes vectorization via image tracing, producing outline vectors. While recent neural approaches [RGLM21] enable end-to-end raster-to-vector conversion, we employ classical techniques for interpretability and real-time performance: *Laplacian smoothing* [Tau95] eliminates pixel-level artifacts in mask boundaries, and a Voronoi-based thinning algorithm [ZS84, AB99] extracts the *medial axis* (centerline) of the segmented path. We then fit cubic Bézier splines using a least-squares error minimization approach. The system automatically detects variable line widths to ensure the subject remains perfectly centered during movement. This geometric abstraction allows the animation to remain fluid regardless of the underlying image resolution.

**Stage 4: Animation Assembly.** The final stage maps the extracted JSON parameters to the host application's internal animation engine (e.g., Adobe InDesign's `MotionPath` API). The system configures keyframes, applies the decoded preset, sets duration, and handles advanced properties such as z-order and transform inheritance. By abstracting the assembly, the pipeline remains platform-agnostic, capable of targeting web-based or desktop-based design environments. The animation is then ready for immediate preview and export.

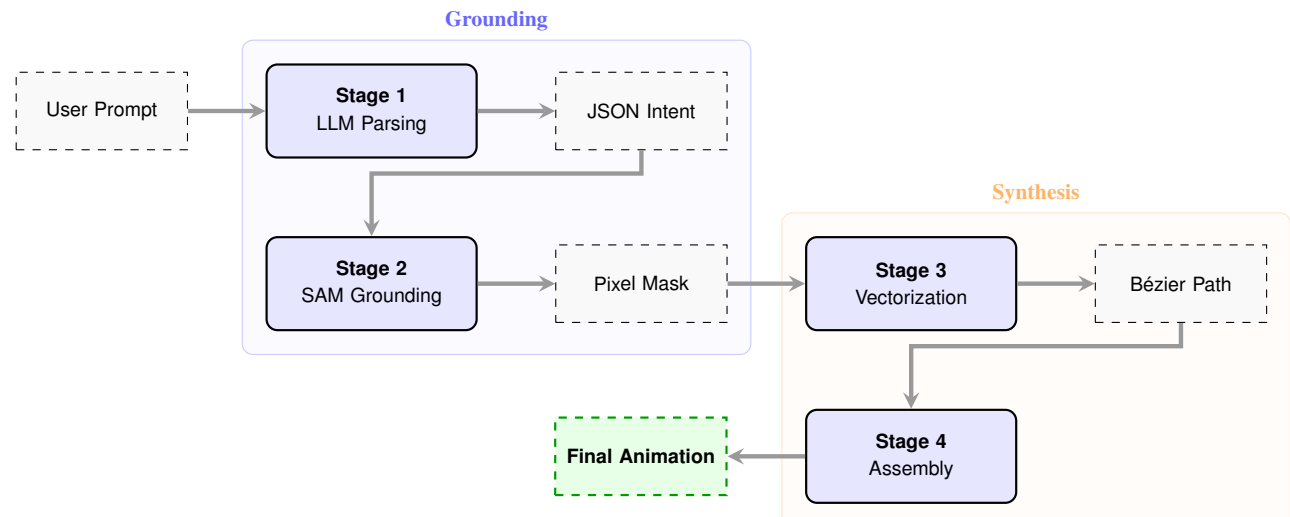

Figure 2: Unified System Architecture: The pipeline translates natural language prompts into structured intent via LLMs, grounds geometry using SAM, synthesizes smooth Bézier paths, and assembles the final animation with environment awareness.

## 4. Pipeline in Action

We demonstrate the pipeline through three scenarios of increasing complexity. The first executes all four stages in their canonical form; subsequent cases show how stages *adapt* to handle depth compositing and 3D transforms.

### 4.1. Baseline: Contour-Following (Mario on Hills)

**Prompt:** "Move Mario along the hilly path."

This case executes the canonical pipeline. **Stage 1** extracts: subject="Mario", entity="hilly path", preset="gallop". **Stage 2** segments the terrain contour. **Stage 3** vectorizes the contour into a smooth Bézier curve. **Stage 4** attaches the path with *gallop* easing, a bouncy timing curve that simulates hopping motion, appropriate for character locomotion. Result: terrain-following animation in under 2 seconds versus 5-10 minutes manually (Figure 3).

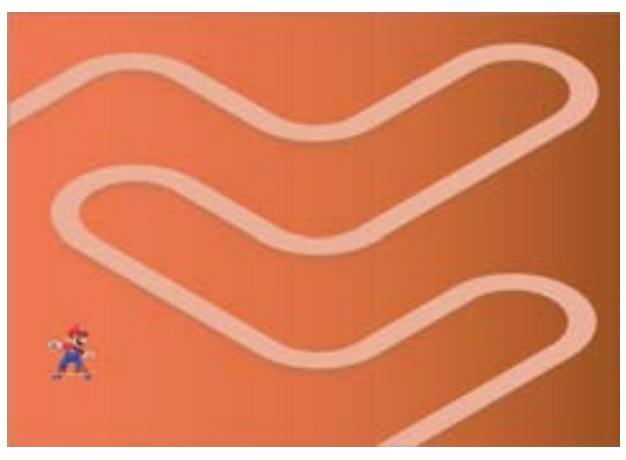

(a) Input: Mario sprite with curvy path

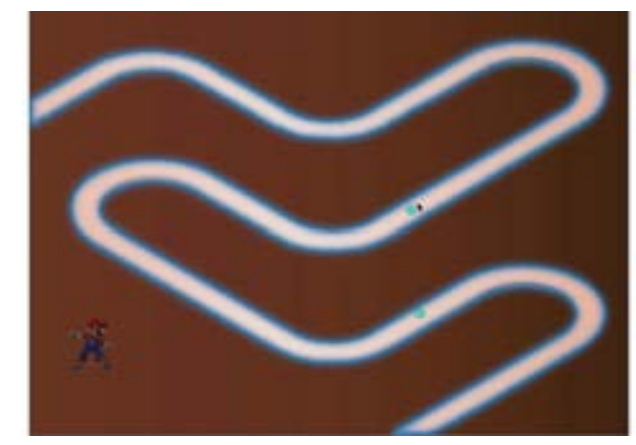

(b) Stage 2: SAM segmentation (blue) with point-click

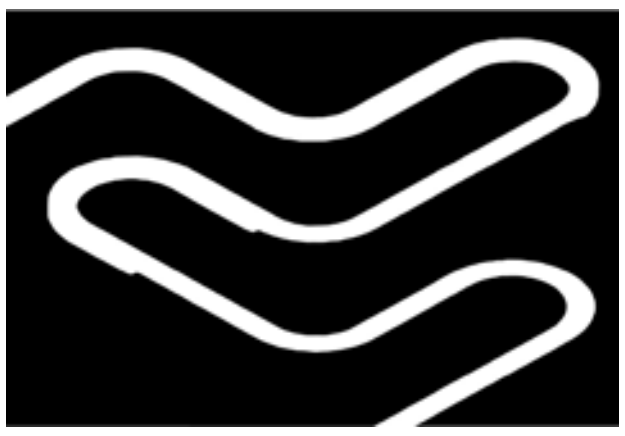

(c) Segmentation mask output

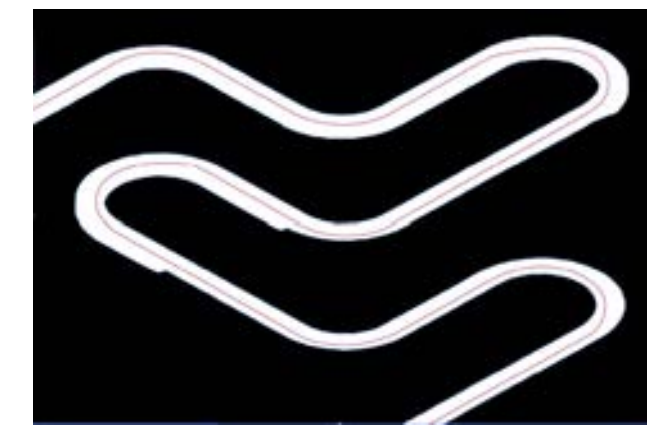

(d) Stage 3: Vectorized motion path (red)

Figure 3: Pipeline execution for "Move Mario along the curvy path." (a) Original artwork with subject and target path. (b) SAM segments the path after user point-click disambiguation. (c) Binary mask extracted from segmentation. (d) Vectorization produces smooth Bézier spline (red centerline) for motion path assignment.

### 4.2. Depth Extension: Orbital Motion (Earth-Moon)

**Prompt:** "Make the Moon orbit around Earth."

**Why variation is needed:** Unlike Mario, the entity here (Earth) is not the motion path itself, but an *obstacle* that creates depth-based occlusion. The Moon must appear in front of Earth for half the orbit and behind it for the remainder. The canonical pipeline has no mechanism for z-order switching mid-animation.

**How stages adapt: Stage 2** segments Earth to establish an *occlusion boundary* rather than a path contour. **Stage 3** *generates* a mathematically defined elliptical path (not extracted from artwork) centered on the reference object. The intersection between this ellipse and the Earth's segmentation mask is used to *split* the path into front and back segments at the mask's boundary coordinates. **Stage 4** produces *two* synchronized animation sequences. The system dynamically assigns a higher z-order to the "front" segment and a lower z-order (behind the reference object) to the "back" segment, creating a seamless pseudo-3D orbital motion in a 2D environment (Figure 4).

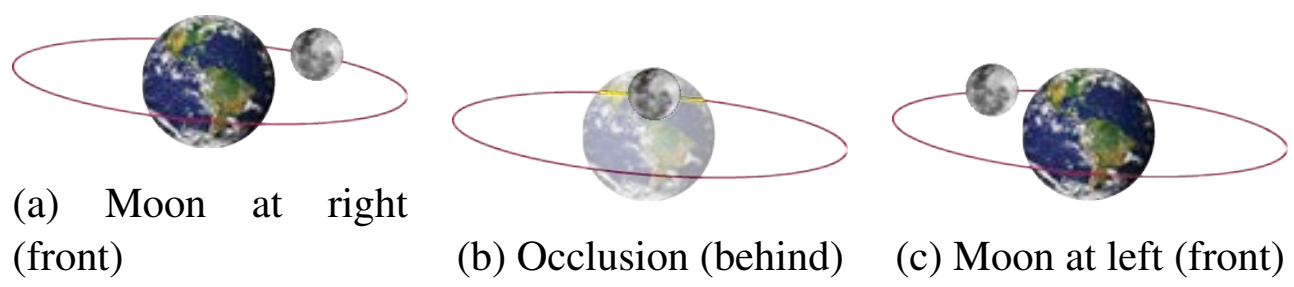

(a) Moon at right (front) (b) Occlusion (behind) (c) Moon at left (front)

Figure 4: Depth Extension: The system segments Earth to establish an occlusion boundary. Keyframes show the Moon (a) in front, (b) occluded behind Earth with the split path highlighted in yellow, and (c) returning to the front, completing the 3D-aware orbit.

### 4.3. 3D Extension: Transform Inheritance (Text Blocks)

**Prompt:** "Fly in The Vision text from the left."

**Why variation is needed:** The text block has a 3D perspective transform applied (rotation on Z-axis). The canonical pipeline generates paths in 2D screen space, but a "fly in from left" animation in 2D would cause the object to slide flatly across the screen rather than moving along its transformed plane. The motion must respect the object's existing 3D orientation.

**How stages adapt: Stage 2** identifies the text block *and* extracts its 3D transformation matrix (stored in the document's metadata). **Stage 3** computes the object's local basis vectors and projects the motion vector into this 3D coordinate space. This ensures the resulting path is aligned with the object's vanishing points. Specifically, if an object is tilted 45° on the Z-axis, the "Fly-In" path is automatically rotated by the same 45° to maintain geometric consistency. **Stage 4** applies this perspective-corrected path, ensuring the motion feels like an extension of the object's inherent 3D property rather than an overlay (Figure 5).

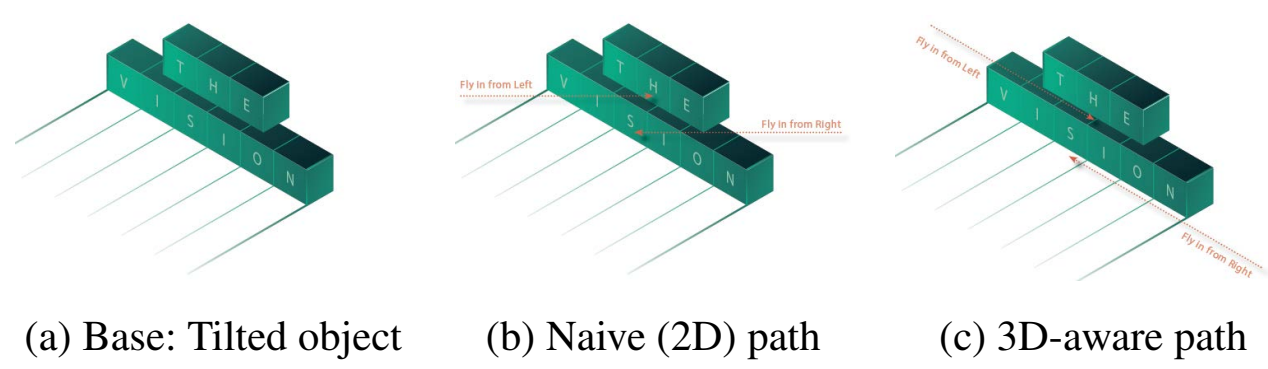

(a) Base: Tilted object (b) Naive (2D) path (c) 3D-aware path

Figure 5: 3D Extension: (a) Input object with existing perspective transform. (b) Naive 2D path synthesis causes flat sliding. (c) Our pipeline projects the motion vector into the object's 3D space, aligning movement with the vanishing point.

## 5. Implementation & Results

**Integration.** The system is implemented as a plugin for Adobe InDesign and Adobe Express, leveraging existing animation engines for keyframe interpolation and playback. The LLM component uses GPT-4 via API; segmentation uses a locally-deployed SAM model for latency optimization.

**Performance.** End-to-end generation (prompt parsing through animation preview) completes in under 2 seconds for typical scenes. The breakdown: LLM inference (~500ms), SAM segmentation (~800ms), vectorization and path synthesis (~400ms), animation assembly (~200ms).

**Supported Presets.** The system recognizes 20+ animation presets including: Appear, Fade In/Out, Fly-In (from all directions), Grow, Shrink, Rotate, Bounce, Dance, Gallop, Pulse, Swoosh, Wave, and Custom path-based animations.

## 6. Discussion

While our system significantly reduces the barrier to entry for motion design, several considerations remain for production environments. First, the dependency on segmentation quality is high. In scenes with low contrast or significant motion blur, SAM may produce noisy boundaries, though our use of Laplacian smoothing mitigates this. Second, the current implementation treats each prompt as an independent event. Handling *multi-turn animation editing*, where a user says “move it faster” or “make the orbit wider”, requires the system to maintain state of the existing animation properties, a feature we are currently developing.

Furthermore, we observe a trade-off between complete automation and creative control. By providing a JSON intent extraction step (Stage 1), we allow power users to manually tweak the animation parameters if the LLM’s interpretation is slightly off. This "glass-box" approach ensures that the system is not just a black-box generator but a collaborative design assistant. Finally, the ability to project 2D motion into 3D spaces suggests that our pipeline could be extended to fully 3D design environments like Adobe Dimension or Aero, where spatial grounding is even more critical.

## 7. Conclusion

We presented Generative Animations, a system that transforms natural language prompts into production-ready, environment-aware animations. By combining LLM-based semantic parsing with SAM-powered visual grounding and robust vectorization, our pipeline bridges the gap between creative intent and geometric execution. The system handles complex scenarios including contour-following paths, depth-aware occlusions, and 3D perspective alignment: capabilities that previously required significant manual effort.

**Limitations & Future Work.** Current limitations include dependency on segmentation quality for complex scenes and the single-entity focus of each animation command. Future work will address **multi-entity interaction**, enabling animations where objects react to each other’s boundaries (e.g., collision avoidance, physics-based bouncing). We also plan to explore **temporal reasoning** for sequencing multiple animations from a single compound prompt. This involves decomposing a multi-action sentence (e.g., “Mario jumps over the hill, then the moon rises”) into a timed sequence of individual pipeline executions with shared state across the animation timeline. Finally, integrating **style inheritance** would allow the LLM to select easing curves that match the visual rhythm of existing document elements.

**Poster:**

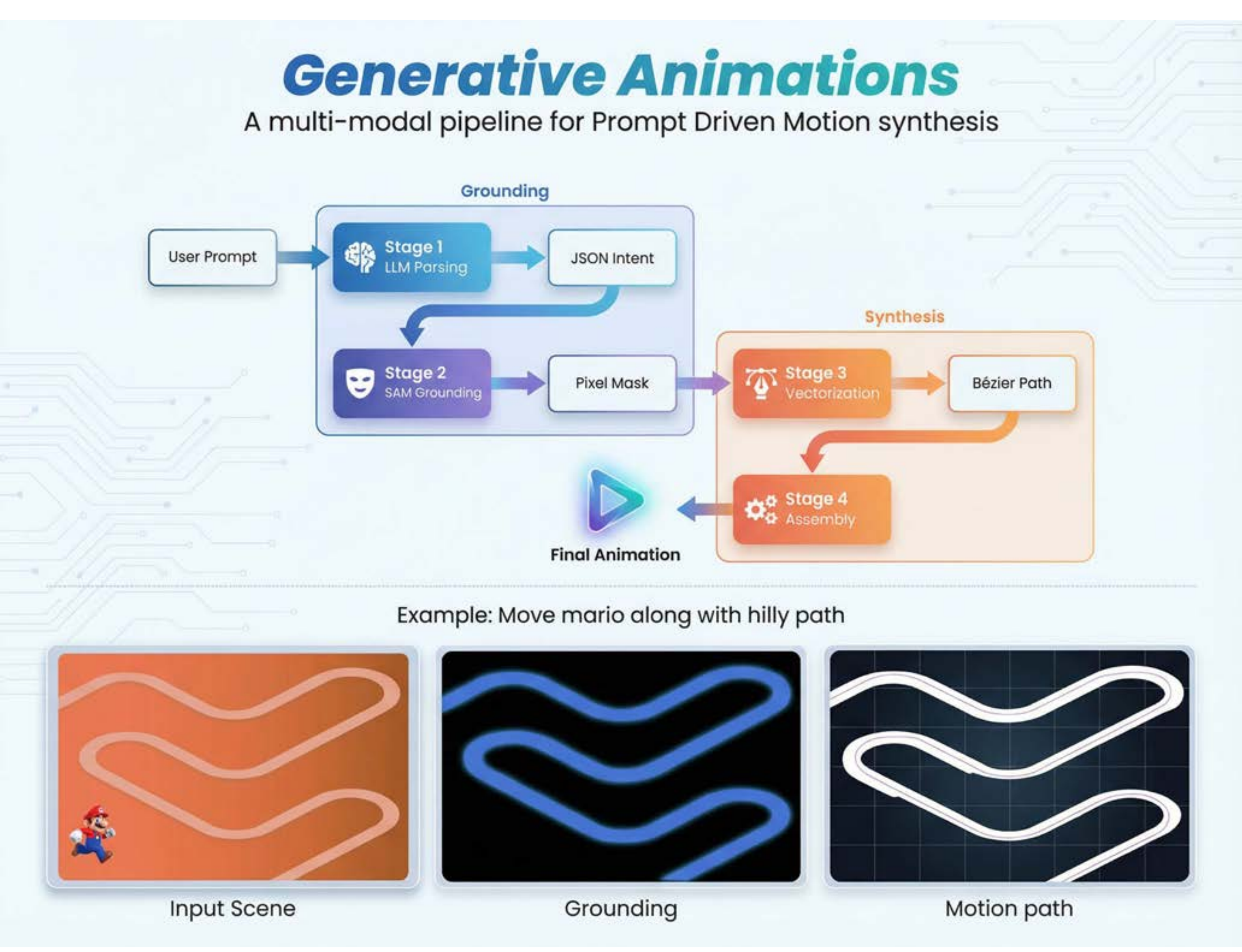
Generative Animations
A multi-modal pipeline for Prompt Driven Motion synthesis
Grounding
User Prompt
Stage 1
LLM Parsing
JSON Intent
Stage 2
SAM Grounding
Pixel Mask
Synthesis
Stage 3
Vectorization
Bézier Path
Stage 4
Assembly
Final Animation
Example: Move mario along with hilly path
Input Scene
Grounding
Motion path